\documentclass[letterpaper, 10 pt, journal, twoside]{ieeetran}

\IEEEoverridecommandlockouts

\usepackage{etextools}
\usepackage{amssymb}
\usepackage{float}
\usepackage{makeidx}
\usepackage{amsmath}
\usepackage{bbm}
\usepackage{epsfig}
\usepackage{epsf}
\usepackage{psfrag}
\usepackage{verbatim}
\usepackage{color}
\usepackage{multirow}
\usepackage{tabularx}
\usepackage{booktabs}
\usepackage[tight,footnotesize]{subfigure}
\usepackage{array}
\usepackage{soul}
\usepackage{footnote}
\usepackage{cite}
\usepackage{dblfloatfix}
\usepackage{tcolorbox}

\usepackage{color, colortbl}
\usepackage[colorlinks,bookmarksnumbered,citecolor=orange,urlcolor=orange]{hyperref}
\usepackage{graphicx}
\graphicspath{{Figures}}
\DeclareGraphicsExtensions{.pdf,.png}
\usepackage{bigstrut}
\usepackage[english]{babel}

\usepackage{csquotes}

\usepackage[T1]{fontenc}
\usepackage{algorithm, setspace}
\usepackage[noend]{algpseudocode}
\usepackage{url}
\usepackage{multirow}
\usepackage{float}
\usepackage{xcolor}
\usepackage{hyperref}
 \hypersetup{
     colorlinks=true,
     linkcolor=orange,
     filecolor=orange,
     citecolor=orange,      
     urlcolor=orange,
     }

\newcommand{\p}[1]{\smallskip \noindent \textbf{{#1}.}}
\newcommand{\eq}[1]{Equation~(\ref{eq:#1})}
\newcommand{\fig}[1]{Figure~\ref{fig:#1}}
\newcommand{\qd}{\hfill\ensuremath{\square}}
\usepackage{balance}
\usepackage{amssymb}

\title{Reward Learning with \\ Intractable Normalizing Functions
}

\author{Joshua Hoegerman and Dylan P. Losey
\thanks{This work is supported in part by NSF Grant \#2222468. \newline The authors are with the Collaborative Robotics Lab (\href{https://collab.me.vt.edu/}{Collab}), Dept. of Mechanical Engineering, Virginia Tech, Blacksburg, VA 24061.
\newline Corresponding author's email: \texttt{jhoegerm@vt.edu}}
}

\begin{document}
\maketitle

\begin{abstract}

Robots can learn to imitate humans by inferring what the human is optimizing for. One common framework for this is Bayesian reward learning, where the robot treats the human's demonstrations and corrections as observations of their underlying reward function. Unfortunately, this inference is \textit{doubly-intractable}: the robot must reason over all the trajectories the person could have provided and all the rewards the person could have in mind. Prior work uses existing robotic tools to approximate this normalizer. In this paper, we group previous approaches into three fundamental classes and analyze the theoretical pros and cons of their approach. We then leverage recent research from the statistics community to introduce \textit{Double MH reward learning}, a Monte Carlo method for asymptotically learning the human's reward in continuous spaces. We extend Double MH to conditionally independent settings (where each human correction is viewed as completely separate) and conditionally dependent environments (where the human's current correction may build on previous inputs). Across simulations and user studies, our proposed approach infers the human's reward parameters more accurately than the alternate approximations when learning from either \textit{demonstrations} or \textit{corrections}. See videos here: \url{https://youtu.be/EkmT3o5K5ko}

\end{abstract}

\begin{keywords}
Intention Recognition,
Learning from Demonstration,
Probabilistic Inference
\end{keywords}


\section{Introduction}

\begin{figure}[t]
	\begin{center}
		\includegraphics[width=1\columnwidth]{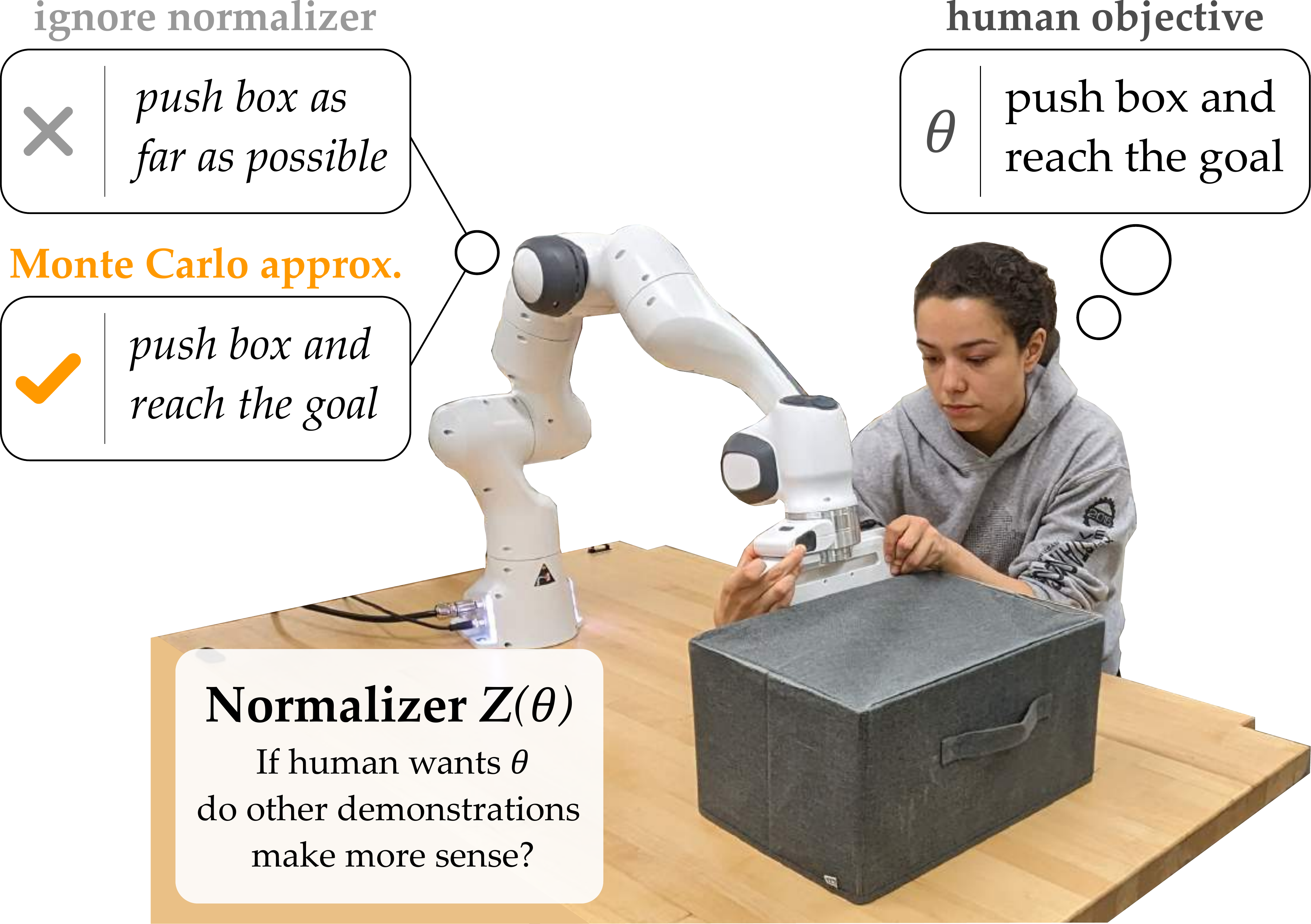}
		\caption{To infer the user's reward, (i.e., objective) robots must compare the human's actual demonstration to other demonstrations the human \textit{could} have provided. Computing this normalizer makes Bayesian reward learning doubly-intractable and ignoring the normalizer entirely can lead to incorrect robot learning. We propose a Monte Carlo method for a more accurate approximation.}
		\label{fig:front}
	\end{center}
	\vspace{-2em}
\end{figure}

Consider a robot arm learning from demonstrations (see \fig{front}). The human guides the robot through example trajectories and the robot tries to infer the human's objective based on their demonstrations. For instance, here the robot arm should learn to slide the box across the table.

State-of-the-art research often tackles this \textit{reward learning} problem using \textit{Bayesian inference} \cite{jeon2020reward, ramachandran2007bayesian, ziebart2008maximum}. The trajectories provided by the human are observations of their latent objective (i.e., their reward function), and the robot recovers a distribution over the rewards by applying Bayes' theorem. Put intuitively: the robot infers rewards under which the human's demonstrations are approximately optimal.

Unfortunately, reward learning in continuous spaces is \textit{doubly intractable}. The robot must normalize across the space of trajectories (i.e., what other demonstrations could the human have provided?), and over the space of rewards (i.e., what else could the human be optimizing for?). Today's robots recognize that they cannot compute these normalizers exactly and so they make a variety of \textit{approximations} \cite{cui2018active, brown2018risk, biyik2022learning, bobu2020quantifying, jonnavittula2021know, kalakrishnan2013learning, finn2016guided, boularias2011relative, li2021learning, hadfield2017inverse, levine2012continuous, dragan2013policy}.

Inaccurate approximations of the normalizing function can lead to incorrect inference: instead of learning what the human meant, the robot learns to perform different and potentially undesirable tasks. Refer back to \fig{front}: here ignoring the normalizer can cause the robot to knock the box off the table. 

In this paper, we focus on inferring the human's reward from demonstrations or corrections. Our insight is that:
\begin{center}\vspace{-0.4em}
\textit{We can enable asymptotic reward learning by leveraging novel Monte Carlo methods from the statistics community.}
\vspace{-0.4em}
\end{center}
Our resulting framework applies to problems where the robot has a predictive model of the environment. Given a sequence of independent or interconnected human demonstrations, we enable robots to accurately learn the human's reward despite doubly intractable normalizing functions. Returning to \fig{front}: our robot learns to push the box correctly using the same amount of data as the baseline. 

Overall, we make the following contributions:

\p{Comparing Approximations} We theoretically and experimentally analyze three existing classes of methods for approximating the normalizer during Bayesian reward learning.

\p{Introducing Double Metropolis-Hastings Sampling} We apply novel statistics research to develop a Double MH algorithm for Bayesian reward learning in continuous spaces.

\p{Learning from Demonstrations and Corrections} Across simulations and user studies, we show that Double MH results in more accurate reward learning from both conditionally independent (e.g., separate) demonstrations and conditionally dependent (e.g., interconnected) corrections.

\section{Related Work} \label{sec:related}

\noindent \textbf{IRL.} Our problem is an instance of Inverse Reinforcement Learning (IRL) where the robot tries to recover the reward that the human wants the robot to optimize for \cite{osa2018algorithmic}. Related work formalizes this as \textit{Bayesian inference} \cite{jeon2020reward, ramachandran2007bayesian, ziebart2008maximum}. Wherein, given a prior distribution and the human's inputs (e.g., state-action pairs \cite{ramachandran2007bayesian} or trajectories \cite{jeon2020reward} provided by the human), the robot arm infers a posterior over the space of possible rewards. In order to connect the human's inputs to rewards, today's robots model the human as a noisily rational teacher \cite{ziebart2008maximum, baker2009action, luce2012individual}. Unfortunately, the human model becomes \textit{intractable} in continuous spaces; without this accurate human model, the robot cannot correctly infer the human's reward.

As we will demonstrate in Section~\ref{sec:problem}, the key challenge is the \textit{normalizing function} of the human model. Within this normalizing function, the robot integrates over the space of all possible human inputs (e.g., their actions or trajectories). This normalizing function makes reward learning doubly intractable: we can use standard Markov chain Monte Carlo (MCMC) methods to eliminate a part of this problem \cite{ramachandran2007bayesian, cui2018active, brown2018risk}, but the normalizing function still remains. Existing IRL algorithms have developed different approaches for dealing with the normalizer. Some works may not explicitly account for the normalizing function \cite{cui2018active, brown2018risk, biyik2022learning}, and there are settings where \textit{ignoring} this normalizer is reasonable. Others use \textit{sampling} to approximate the normalizer: this includes sampling uniformly from the trajectory space \cite{bobu2020quantifying, jonnavittula2021know}, sampling trajectories close to the human's inputs \cite{kalakrishnan2013learning}, or using importance sampling to convert between the robot's trajectory distribution and a uniform distribution over trajectories \cite{finn2016guided, boularias2011relative}. Finally, related works have also used the \textit{maximum} trajectory reward in place of the normalizing function \cite{li2021learning}. This substitution can be justified using Laplace's approximation \cite{levine2012continuous, dragan2013policy}.

To summarize, prior work on Bayesian IRL uses ignore, sampling, and maximum approaches to estimate the normalizing function and infer the human's reward. In this paper, we theoretically and experimentally compare these different approaches to understand their relative advantages.

\p{Approximate Bayesian Inference} Within the robotics community, we have developed a variety of techniques for learning with an intractable normalizing function. But what about work \textit{outside} of robotics? Statistics research has recently proposed several Markov chain Monte Carlo (MCMC) algorithms for Bayesian inference in the presence of intractable normalizing functions \cite{liang2010double, lyne2015russian, alquier2016noisy, park2018bayesian}. These approaches modify techniques such as Metropolis-Hastings sampling to obtain computationally efficient and accurate algorithms for inferring \textit{doubly intractable} posterior distributions. In this paper, we apply recent breakthroughs from the statistics community to propose a new method for Bayesian reward learning.

\vspace{-1em}
\section{Problem Formulation} \label{sec:problem}

We consider settings where a robot arm is using Bayesian inference to learn from human examples. The human teacher knows what task the robot should perform. More specifically, the human has in mind a reward function that the robot should optimize, and the robot is trying to infer this reward from the human's data. The human might provide complete examples of their desired behavior (i.e., demonstrations), or they could just modify snippets of the robot's existing motion (i.e., corrections). We recognize that humans are not perfect teachers: when showing the robot how to carry a glass of water, the human may not have enough time or ability to meticulously orchestrate every joint. The robot views the human as a \textit{noisily rational} agent that approximately maximizes their reward.

\p{Task and Reward} This problem is an instance of a Markov decision process (MDP) where the robot does not know the reward function. Let the MDP be a tuple $M = \langle \mathcal{S}, \mathcal{A}, f, r, T\rangle$ where $s \in \mathcal{S}$ is the system state and $a \in \mathcal{A}$ is the robot's action. For example, $s$ could be the arm's joint position and the pose of the cup, and $a$ could be the robot's joint velocity. At timestep $t$, the system transitions to a new state according to the deterministic dynamics $s^{t+1} = f(s^t, a^t)$. The task ends after a total of $T$ timesteps. Let $\xi = (s^0, \ldots s^T)$ be the robot's \textit{trajectory}, i.e., the sequence of visited states across $T$ timesteps, and let $\Xi$ be the set of possible trajectories.

During every timestep, the robot receives a scalar reward $r(s)$. Remember that the robot \textit{does not} know the desired reward function. Without loss of generality, we will write the reward as $r(s, \theta)$, where vector $\theta \in \mathbb{R}^d$ captures the aspects of the reward function that the robot does not know. For example, in related works \cite{jeon2020reward, osa2018algorithmic} the reward function is often a linear combination of features such that $r(s, \theta) = \theta \cdot \phi(s)$. Here $\phi : \mathcal{S} \rightarrow \mathbb{R}^d$ is a $d$-length feature vector that captures task-relevant aspects of the state (e.g., the distance from the table, the orientation of the cup), and $\theta$ determines the relative importance of these features. Across an entire trajectory, the robot's cumulative reward is: $R(\xi, \theta) = \sum_{s \in \xi} r(s, \theta)$. If the reward function is a linear combination of features, this simplifies to: $R(\xi, \theta) = \theta \cdot \sum_{s \in \xi} \phi(s) = \theta \cdot \Phi(\xi)$.

\p{Human Data} The robot is attempting to learn the reward (and more precisely, the unknown parameters $\theta$) from human examples. Let $\mathcal{D} = \{\xi_1, \ldots \xi_K\}$ be a dataset of $K$ trajectories provided by the human expert. Our approach is not tied to any specific way of gathering this dataset. The trajectories could be collected \textit{offline} as the human kinesthetically guides the robot through task demonstrations \cite{kalakrishnan2013learning, finn2016guided, biyik2022learning}. Although, we could also add improved trajectories \textit{online} as the human corrects the robot arm \cite{losey2022physical, li2021learning, bobu2020quantifying}. In either case, we aggregate the human's trajectories into the dataset $\mathcal{D}$.

For simplicity, we now assume that the human teacher only provides a \textit{single} trajectory, i.e., $\mathcal{D} = \xi$. In Section~\ref{sec:mcmc} we will extend our analysis to a dataset of $K$ trajectories.

\p{Bayesian Inference} The robot infers the parameters $\theta$ from the human's trajectory $\xi$. Let $P(\theta \mid \xi)$ denote the probability that the human is optimizing for reward parameters $\theta$ given the input trajectory $\xi$. Applying Bayes' theorem:
\begin{equation} \label{eq:P1}
    P(\theta \mid \xi) \propto P(\xi \mid \theta) \cdot P(\theta)
\end{equation}
where $P(\theta)$ is the prior and $P(\xi \mid \theta)$ is the likelihood function. Intuitively, $P(\xi \mid \theta)$ expresses how likely it is (from the robot's perspective) that the human provides trajectory $\xi$ given the human is optimizing for reward parameters $\theta$.

\p{Human Model} The likelihood function $P(\xi \mid \theta)$ is a human model: it tries to capture how the human teacher maps their hidden objective to an example trajectory. Prior work in behavioral economics \cite{luce2012individual}, cognitive science \cite{baker2009action}, and reward learning \cite{ziebart2008maximum} suggests that humans are \textit{noisily rational} agents. These humans are not perfect: instead of always choosing the best possible trajectory, noisily rational humans are exponentially more likely to select behaviors with higher rewards. Under the noisily rational model: 
\begin{equation} \label{eq:P2}
    P(\xi \mid \theta) = \frac{\exp \big( \beta \cdot R(\xi, \theta) \big)}{\int_{\Xi} ~\exp \big( \beta \cdot R(\xi', \theta) \big) ~d\xi'}
\end{equation}
Where $\beta \in [0, \infty)$ is a hyperparameter set by the designer. As $\beta \rightarrow 0$ each trajectory becomes equally likely and the robot treats the human as a random agent. At the other extreme, as $\beta \rightarrow \infty$ the human is only likely to input optimal trajectories and the robot views the human as perfectly rational.

\p{Normalizing Function} The numerator of \eq{P2} is straightforward: we simply substitute $\xi$ and $\theta$ into the reward function and evaluate. But once we find $\exp \big( \beta \cdot R(\xi, \theta) \big)$, how good (i.e., how likely) is that trajectory? We need a sense of scale to understand the relative reward for $\xi$ as compared to the alternatives --- perhaps there is another trajectory $\xi'$ that achieves a much higher reward. This is where the denominator of \eq{P2} comes in. The denominator is a \textit{normalizing function} that integrates over the continuous space of possible trajectories $\xi' \in \Xi$ given reward parameters $\theta$. We refer to the normalizing function as $Z(\theta)$:
\begin{equation} \label{eq:P3}
    Z(\theta) = \int_{\Xi} ~\exp\big( \beta \cdot R(\xi', \theta) \big) ~d\xi'
\end{equation}
The normalizing function $Z(\theta)$ serves to calibrate our human model. Importantly, $Z(\theta)$ can be different for different values of $\theta$. We will show examples of how $Z$ changes (or does not change) as a function of $\theta$ in the following sections. 

\p{Summary} To infer the human's reward through Bayesian inference we need to find $P(\xi \mid \theta)$ in \eq{P1}. But to get $P(\xi \mid \theta)$ we first must be able to solve \eq{P3}, and this normalizing function is \textit{intractable} when $\Xi$ is a continuous manifold \cite{finn2016guided}. This leads to our core problem: how should robots approximate (or avoid) the normalizing function $Z(\theta)$ when learning from human data?
\vspace{-0.5em}
\section{Approximating the Normalizer} \label{sec:norm}

In this section, we analyze three state-of-the-art approaches for dealing with the normalizing function in Bayesian inference. In Section~\ref{sec:M1} we prove that robots can completely ignore the normalizing function if $Z$ does not depend on $\theta$; we also identify the necessary conditions for this special case when the reward is a linear combination of features. Moving beyond this special case, we next explore two methods for approximating $Z(\theta)$. In Section~\ref{sec:M2} we discuss a sampling approach and in Section~\ref{sec:M3} we use the maximum value in place of the normalizer. We prove that the maximum approach will match or outperform the sampling approach as $\beta \rightarrow \infty$ in our noisily rational human model.

\p{Working Example} We first introduce a simplified example to illustrate the analysis throughout this section. Consider a robot arm that is learning how to carry a cup. The robot can hold the cup at any angle between $0$ radians (horizontal) and $\pi/2$ radians (vertical). The human teacher inputs a trajectory $\xi = s$ where they specify a single orientation of the cup. The reward is: $r(s, \theta) = -5\cos(\theta) \cdot (s+1) -\sin(\theta) \cdot (\pi/2 - s)$. Based on the human's input $\xi$, the robot is trying to determine whether (a) $\theta = 0$ and the robot should hold the cup horizontally at angle $s=0$, or whether (b) $\theta = \pi/2$ and the robot should hold the cup vertically at angle $\pi/2$. Let the robot have a uniform prior over these two possible reward parameters. Applying Equations~(\ref{eq:P1})-(\ref{eq:P3}), the robot's belief that $\theta = 0$ is:
\begin{equation} \label{eq:M1}
    P(0 \mid \xi) = \frac{\exp\big( \beta R(\xi, 0)\big)}{\exp\big( \beta R(\xi, 0)\big) + \frac{Z(0)}{Z(\pi/2)} \cdot \exp\big( \beta R(\xi, \frac{\pi}{2})\big)}
\end{equation}
In this simple example, numerical integration is possible and we can exactly find $Z(0)$ and $Z(\pi/2)$. Plugging these exact values into \eq{M1} gives us the \textit{ideal} belief. Put another way, this is what the robot \textit{should} learn. We will compare this ideal result to approaches that approximate the normalizer.

\vspace{-1em}
\subsection{Ignoring the Normalizing Function} \label{sec:M1}

In some settings, it may be reasonable to ignore the normalizing function altogether. Methods such as \cite{cui2018active, brown2018risk, biyik2022learning} use MCMC sampling to cancel out the partition function $P(\xi)$, but they do not explicitly account for the normalizing function within $P(\xi \mid \theta)$. In our working example for instance, these approaches may omit the $Z$ terms from \eq{M1}.

\p{Proposition 1} \textit{We can ignore the normalizing function when $Z$ does not depend on $\theta$}.

\p{Proof} Consider a problem setting where $Z$ is a constant, i.e., where $Z(\theta_i) = Z(\theta_j)$ for any choice of $\theta_i$ and $\theta_j$. When we substitute \eq{P2} back into \eq{P1} to infer the reward, both the numerator and denominator are multiplied by $Z$ and this normalizing constant cancels out:
\begin{equation}
    P(\theta \mid \xi) = \frac{Z}{Z} \Bigg(\frac{\exp\big( \beta \cdot R(\xi', \theta)\big) \cdot  P(\theta)}{\int_{\Theta}~ \exp\big( \beta \cdot R(\xi, \theta') \big) \cdot  P(\theta') ~ d\theta'}\Bigg)
\end{equation}
Looking specifically at the working example in \eq{M1}, if $Z(0) = Z(\pi/2)$ then the $Z$ terms cancel. \qd

\smallskip

So far our analysis shows that we can ignore $Z$ without any loss in performance \textit{if} the normalizer is a constant. But when is this the case? To answer this question we focus on a common framework where the reward function is a linear combination of features, $R(\xi, \theta) = \theta \cdot \Phi(\xi)$. We find that:

\p{Proposition 2} \textit{If $R(\xi, \theta) = \theta \cdot \Phi(\xi)$ and $\theta \in \mathbb{R}^d$ is a $d$-dimensional unit vector, $Z$ does not depend on $\theta$ \emph{if and only if} the feature space $\Phi(\Xi)$ is a sphere centered at zero with radius $\sigma \geq 0$, i.e., $\Phi(\Xi) = \{v \in \mathbb{R}^d : \| v\| = \sigma\}$}.

\p{Proof} Let $\theta_i$ and $\theta_j$ be two arbitrary unit vectors. Consider \eq{P2} with $\beta \in [0, \infty)$. For the integrals $Z(\theta_i)$ and $Z(\theta_j)$ to be equal, for every $\xi \in \Xi$ there must exist some $\xi' \in \Xi$ such that: $\theta_i \cdot \Phi(\xi) = \theta_j \cdot \Phi(\xi')$. This is only satisfied when $\Phi(\Xi)$ is proportional to a unit sphere. \qd

\smallskip

In practice, it is challenging to ensure the feature space is a sphere. Not only do we need the average of each individual feature to be zero, but the combination of features must always have the same radius. Consider \fig{front} where the human is teaching the robot arm: along the human's trajectory they might minimize the robot's height and orientation, resulting in a feature vector $\Phi(\xi)$ where each element is close to zero. Alternatively, the human might move the robot far from the table while changing the orientation, leading to a $\Phi(\xi)$ where each element is close to one. The magnitude of these two feature vectors is different --- and thus we \textit{cannot} apply Proposition 2 and ignore the normalizer.

\begin{figure}[t]
	\begin{center}
		\includegraphics[width=1\columnwidth]{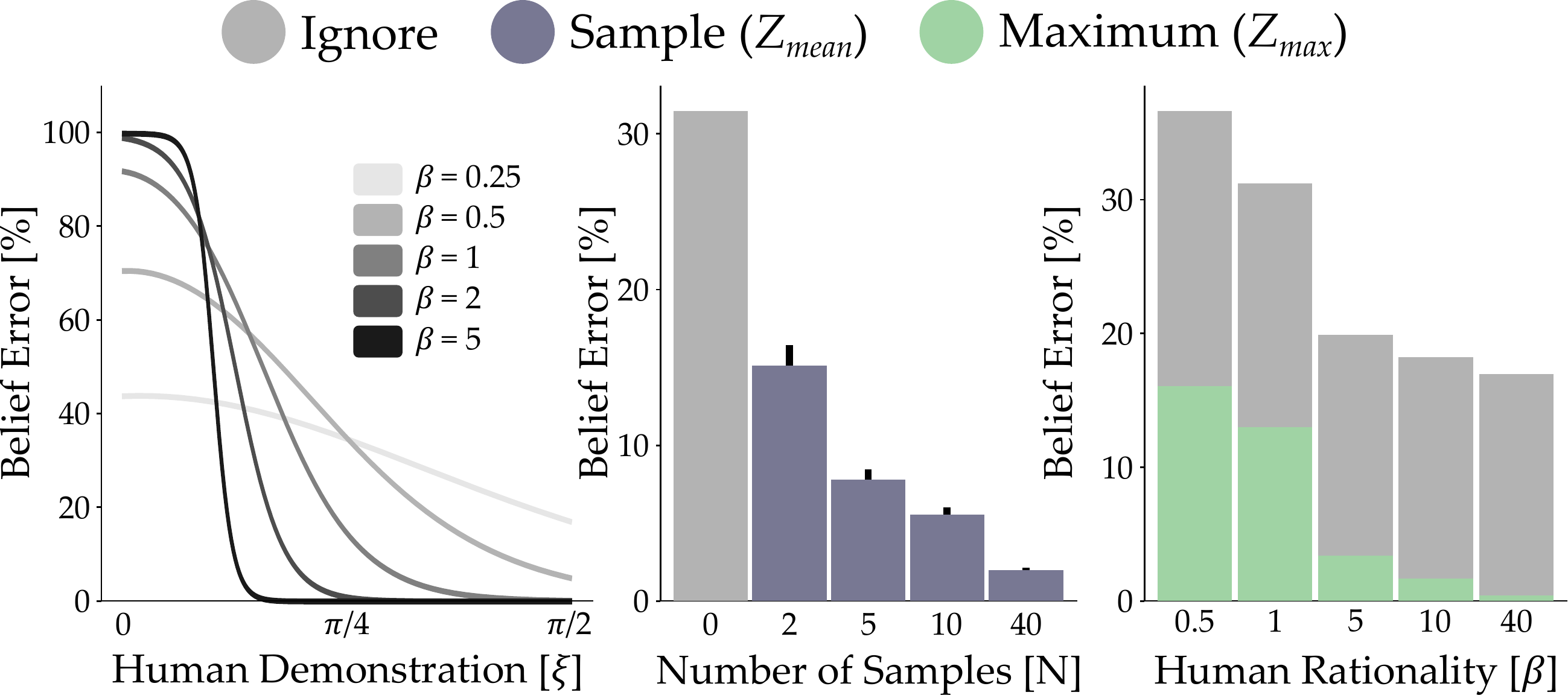}
		\caption{Approximating the normalizer in our working example. \textit{Belief Error} is the difference between evaluating \eq{M1} with the exact $Z(\theta)$ and \eq{M1} with approximations for $Z(\theta)$. The lower error indicates the robot learned the correct belief. (Left) The human provides demonstrations $\xi$ of carrying the cup at different angles. If we \textbf{ignore} the normalizer, as $\beta \rightarrow \infty$ the robot always learns to carry the cup vertically, even when the human wants the opposite. (Middle) Here we left $\beta = 1.0$. As the number of \textbf{samples} for $Z_{mean}$ increases, the belief error converges to zero. (Right) As $\beta \rightarrow \infty$ the error with the \textbf{maximum} approach converges to zero.}
		\label{fig:ex1}
	\end{center}
	\vspace{-2em}
\end{figure}

In Propositions 1 and 2 we have identified a special case where the robot does not need to evaluate $Z$. However, for common settings where $Z$ \textit{is} a function of $\theta$, ignoring the normalizing function results in errors in the robot's learning. See our working example in \fig{ex1} where we plot the error between \eq{M1} with and without $Z(\theta)$.

\vspace{-1em}
\subsection{Approximating the Normalizer with Sampling} \label{sec:M2}

Instead of ignoring the normalizing function, next we will estimate $Z(\theta)$. One approach is to approximate the integral in \eq{P3} through sampling \cite{bobu2020quantifying, jonnavittula2021know, kalakrishnan2013learning, finn2016guided, boularias2011relative, hadfield2017inverse}. Let $\{\xi_1, \ldots, \xi_N\}$ be $N$ trajectories sampled uniformly at random from the trajectory space $\Xi$. We use these samples to approximate the mean value of $\exp\big( \beta \cdot R(\xi, \theta) \big)$ as shown below: 
\begin{equation} \label{eq:M3}
    Z_{mean}(\theta) = \frac{1}{N} \sum_{i = 1}^N ~\exp\big( \beta \cdot R(\xi_i, \theta) \big)
\end{equation}
Applying the law of the unconscious statistician, this estimate noisily converges to the actual mean as the number of $\xi$ samples increases. It may seem unintuitive at first that we are estimating the mean and not $Z(\theta)$. However, $Z(\theta)$ is equal to this mean multiplied by a volumetric constant that does not depend on $\theta$; because the constant cancels out during Bayesian inference, we only need the mean.

We test this sampling approach on our working example in \fig{ex1}. Compared to a robot that ignores the normalizing function, attempting to estimate $Z(\theta)$ leads to more accurate learning. But we do recognize that the sampling approach comes with an assumption: specifically, we now assume that the robot knows the space of possible trajectories $\Xi$.

\vspace{-1em}
\subsection{Approximating the Normalizer as the Maximum} \label{sec:M3}

Other state-of-the-art algorithms use a maximum value in place of the normalizing function \cite{li2021learning, levine2012continuous, dragan2013policy}. These approaches find the maximum of the numerator in \eq{P2}, and then treat this maximum as the denominator:
\begin{equation} \label{eq:M4}
    Z_{max}(\theta) = \max_{\xi \in \Xi} ~ \exp \big( \beta \cdot R(\xi, \theta) \big)
\end{equation}
Intuitively, scaling by $Z_{max}$ makes sense because it ensures that the resulting $P(\theta \mid \xi)$ is always less than or equal to one. Recall that for the sampling approach $Z_{mean}$ converges as the number of $\xi$ samples increases. Interestingly, we find an analogous convergence for the maximum approach:

\p{Proposition 3} \textit{The error between Bayesian inference with $Z_{max}$ and an ideal robot that uses $Z$ converges to zero as the human becomes increasingly optimal, i.e., as $\beta \rightarrow \infty$.}

\p{Proof} For any given $\theta$, let there be a single trajectory $\xi^*$ that maximizes the human's reward such that $R(\xi^*, \theta) > R(\xi, \theta)$ for all $\xi \in \Xi \setminus \xi^*$. Use numerical integration to estimate $Z$:
\begin{equation} \label{eq:M5}
    Z(\theta) \doteq  C \bigg(Z_{max}(\theta) + \sum_{i = 1}^N~\exp\big( \beta \cdot R(\xi_i, \theta) \big) \bigg)
\end{equation}
where $C$ is a volumetric constant that cancels out in Bayesian inference, and $\{\xi_1, \ldots, \xi_N\}$ are $N$ non-optimal trajectories sampled uniformly at random from $\Xi \setminus \xi^*$. Taking the limit as $\beta \rightarrow \infty$, and remembering that $R(\xi^*, \theta) > R(\xi, \theta)$, we have that $Z_{max}(\theta)$ dominates the remaining terms. Accordingly, as $\beta \rightarrow \infty$ \eq{M5} converges to $C \cdot Z_{max}(\theta)$, and the difference between a robot that learns using $Z(\theta)$ and a robot that learns using $Z_{max}$ converges to zero. \qd

\smallskip

We apply Proposition 3 to our working example in \fig{ex1}. As expected, using $Z_{max}$ as the normalizing function becomes increasingly accurate as $\beta \rightarrow \infty$. But now that we have two different methods for approximating the normalizer, we are left with a decision: when should designers use $Z_{mean}$ and when should designers use $Z_{max}$? The answer to this question varies as the problem setting and reward function change. However, we do find a general trend:

\p{Proposition 4} \textit{As $\beta \rightarrow \infty$ in the human model, robots learn an equal or more accurate estimate of $P(\theta \mid \xi)$ using Bayesian inference with $Z_{max}$ instead of $Z_{mean}$.}

\p{Proof} From Proposition 3 we already know that $Z_{max}$ converges to ideal performance as $\beta\rightarrow \infty$. It only remains to evaluate the performance of $Z_{mean}$. Recall from \eq{M3} that $Z_{mean}$ samples $N$ trajectories from space $\Xi$. Because $N$ is a finite number, there will be cases when the robot does not sample the optimal trajectory $\xi^*$. Compare the numerical integration in \eq{M5} to the sampled mean in \eq{M3}. If the robot does not sample $\xi^*$ in \eq{M3}, then as $\beta \rightarrow \infty$ \eq{M3} is not necessarily proportional to \eq{M5}, where $Z$ is dominated by the exponentiated reward of $\xi^*$. Because $Z_{mean}$ is not necessarily proportional to $Z$, a robot that learns using $Z_{mean}$ may not match the performance of an ideal robot that learns with $Z$. \qd

\begin{figure}[t]
	\begin{center}
		\includegraphics[width=1\columnwidth]{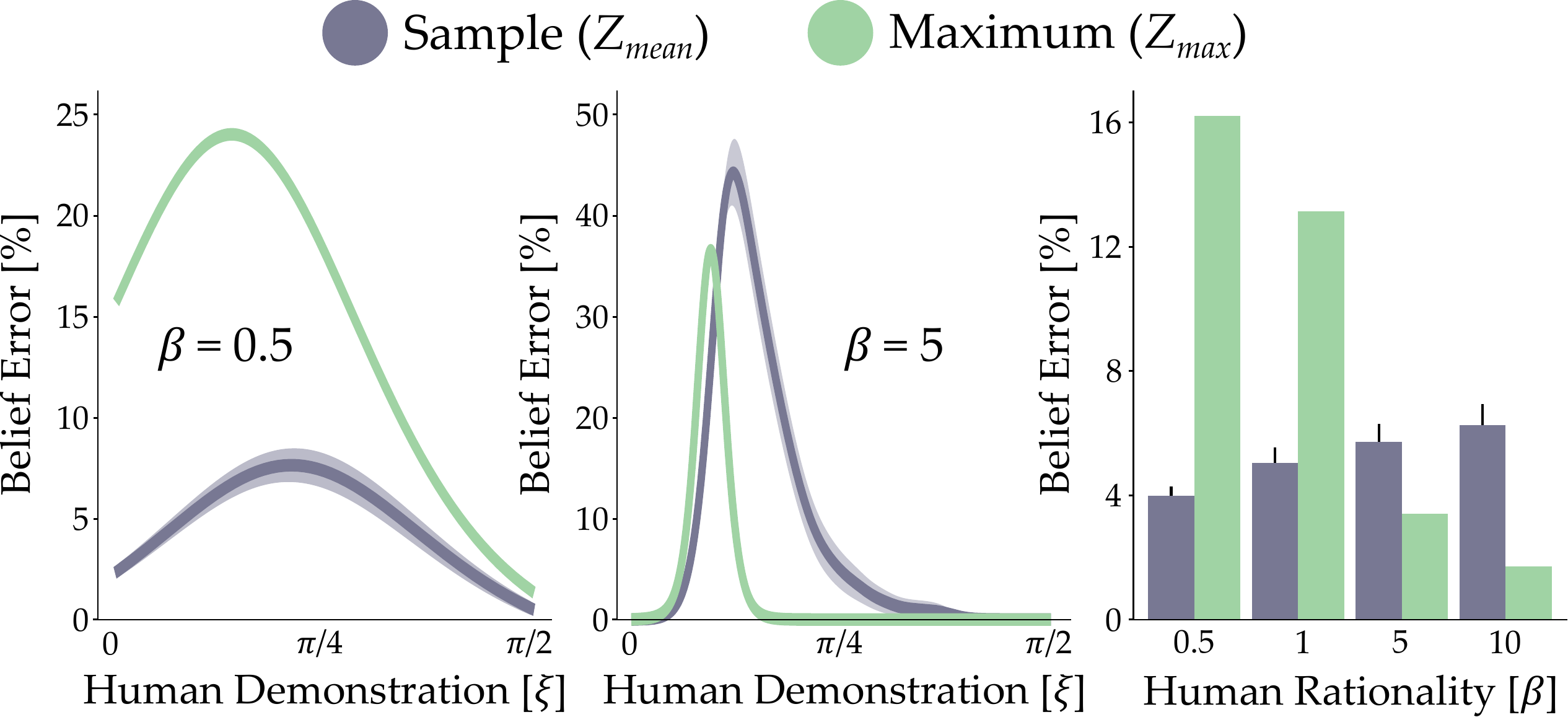}
		\caption{Comparing \textbf{sampling} and \textbf{maximum} approaches in our working example. Remember that $\beta$ from \eq{P2} captures how close-to-optimal the human is. (Left) Error when $\beta = 0.5$. (Middle) Error when $\beta = 5$. (Right) For lower values of $\beta$ we find that the sampling approach is more accurate. However, as $\beta\rightarrow\infty$ the maximum approach leads to less error. For \textbf{sampling} we perform $100$ separate runs where $Z_{mean}$ samples $N=10$ trajectories each run; the shaded region and bars show standard error.}
		\label{fig:ex2}
	\end{center}
	\vspace{-2em}
\end{figure}

\smallskip

We demonstrate Proposition 4 in \fig{ex2}. For \textit{lower values} of $\beta$ we find that approximating the normalizer with \textit{sampling} outperforms the maximum approach. By contrast, for \textit{higher values} of $\beta$ using the \textit{maximum} as the normalizing function results in more accurate learning. The exact trade-off point is problem-specific, but this general trend holds across our theoretical analysis and experimental results.
\vspace{-1em}
\section{Scaling up with \\Metropolis-Hastings Sampling} \label{sec:mcmc}

In Section~\ref{sec:norm}, we analyzed the normalizing function when the human only provides a single trajectory, i.e., when $\mathcal{D} = \xi$. In this section, we scale up to general cases where the robot is learning from a dataset $\mathcal{D}$ of $K$ trajectories. As we scale up, we recognize that the space of rewards $\Theta$ is \textit{continuous}. In our working example we assumed that the human wanted the robot to hold the cup either horizontally or vertically; but, more generally, the human may want the robot to hold the cup at any angle. To enable Bayesian inference in continuous reward spaces, we turn to \textit{Metropolis-Hastings (MH) sampling} \cite{russell2022artificial}. We first formulate conditionally independent and dependent reward learning from multiple trajectories (Section~\ref{sec:A1}) and then combine approaches for approximating the normalizer with MH sampling (Section~\ref{sec:A2}). In Section~\ref{sec:A3} we introduce the \textit{Double MH algorithm} for Bayesian reward learning.

\vspace{-1em}
\subsection{Learning from Multiple Trajectories} \label{sec:A1}

Let $\mathcal{D} = \{\xi_1, \ldots, \xi_K\}$ be a dataset of $K$ trajectories input by the human teacher. The probability that the human has in mind reward $\theta$ given dataset $\mathcal{D}$ is:
\begin{equation} \label{eq:A1}
    P(\theta \mid \mathcal{D}) \propto P(\mathcal{D} \mid \theta) \cdot P(\theta)
\end{equation}
Similar to \eq{P1}, here $P(\theta)$ is the prior over the space of rewards and $P(\mathcal{D} \mid \theta)$ is the likelihood the human inputs $\mathcal{D}$ given their reward is parameterized by $\theta$.

\p{Conditionally Independent} If the human selects each trajectory separately then the human's inputs are conditionally \textit{independent}. For instance, consider a human that repeatedly demonstrates a task to the robot: each demonstration depends on their reward $\theta$, but the human does not reason over $\xi_i$ when selecting $\xi_j$ \cite{biyik2022learning, ziebart2008maximum, jeon2020reward}. When the trajectories are conditionally independent \eq{A1} reduces to:
\begin{equation} \label{eq:A2}
    P(\theta \mid \mathcal{D}) \propto P(\theta) \cdot \prod_{i = 1}^K P(\xi_i \mid \theta)
\end{equation}
Plugging in our human model from \eq{P2}, we reach:
\begin{equation} \label{eq:A3}
    P(\theta \mid \mathcal{D}) \propto \frac{\exp \Big( \beta \cdot \sum_{i = 1}^K R(\xi_i, \theta) \Big) \cdot P(\theta)}{Z(\theta)^K}
\end{equation}
where $R(\xi, \theta) = \sum_{s \in \xi} r(s, \theta)$ is the total reward along input $\xi$ and $Z(\theta)$ is the normalizing function from \eq{P3}.

\p{Conditionally Dependent} Alternatively, if the human provides multiple interconnected trajectories the human's inputs are conditionally \textit{dependent}. For example, imagine a human that iteratively improves the robot's motion by making small corrections: the human's input $\xi_j$ will depend on the human's reward but also on the distance between $\xi_j$ and the previous trajectory $\xi_i$ \cite{li2021learning}. In this case, we cannot simplify \eq{A1}. Instead, we define an augmented human model:
\begin{equation} \label{eq:A4}
    P(\mathcal{D} \mid \theta) = \frac{\exp \big( \beta \cdot \mathbf{R}(\mathcal{D}, \theta) \big)}{\int_{\mathbb{D}} ~\exp \big( \beta \cdot \mathbf{R}(\mathcal{D}', \theta) \big) ~d\mathcal{D}'}
\end{equation}
where $\mathbf{R}$ is the total reward over dataset $\mathcal{D}$. Going back to our example of a human that makes small improvements, $\mathbf{R}$ could be \cite{li2021learning}: $\mathbf{R}(\mathcal{D}, \theta) = \sum_{i = 2}^K R(\xi_i, \theta) - \| \xi_i - \xi_{i-1}\|^2$. Looking at the denominator of \eq{A4}, for conditionally dependent trajectories we need to normalize over the entire dataset $\mathcal{D}$ rather than the individual inputs $\xi$:
\begin{equation} \label{eq:A5}
    \mathbf{Z}(\theta) = \int_{\mathbb{D}} ~\exp \big( \beta \cdot \mathbf{R}(\mathcal{D}', \theta) \big) ~d\mathcal{D}' 
\end{equation}
Here $\mathbb{D}$ is the space of possible datasets $\mathcal{D}$. To find $P(\theta \mid \mathcal{D})$ and perform Bayesian inference we plug \eq{A4} with normalizing function $\mathbf{Z}(\theta)$ back into \eq{A1}.

\vspace{-1em}
\subsection{Metropolis-Hastings Sampling} \label{sec:A2}

Regardless of whether the human's inputs are conditionally independent or conditionally dependent, we want to use dataset $\mathcal{D}$ to infer the reward parameters $\theta$. This leads us back to the posterior distribution $P(\theta \mid \mathcal{D})$. To evaluate $P(\theta \mid \mathcal{D})$ in \eq{A1} we have to deal with \textit{another} normalizer; specifically, the denominator $P(\mathcal{D}) = \int_{\Theta} P(\mathcal{D} \mid \theta) \cdot P(\theta) d\theta$. When $\Theta$ is a discrete space (e.g., in our working example where the human wants the cup either horizontal or vertical) we can compute this denominator and find the probability of each $\theta \in \Theta$. But when $\Theta$ is continuous, Bayesian inference becomes \textit{doubly intractable} and we cannot typically find closed-form expressions for $P(\theta \mid \mathcal{D})$. Instead, the robot learner uses the MH algorithm to \textit{sample} values of $\theta$ from the non-normalized posterior, i.e., $\theta \sim P(~ \cdot \mid \mathcal{D})$.

\begin{algorithm}[t]
    \setstretch{1.0}
    \caption{Bayesian Reward Learning \\ with Normalizer Approximation}
    \label{alg:norm}
    \begin{algorithmic}[1]
    \State $\theta \gets$ sample from $P(\theta)$
    \For{each iteration}
        \State $\theta' \gets$ sample from $\Theta$ near $\theta$
        \State \textit{Conditionally Independent:} \smallskip
        \State $\frac{P(\theta' \mid \mathcal{D})}{P(\theta \mid \mathcal{D})} \gets \frac{\exp \big( \beta \cdot \sum_{i = 1}^K R(\xi_i, \theta') \big) \cdot Z(\theta)^K \cdot P(\theta')}{\exp \big( \beta \cdot \sum_{i = 1}^K R(\xi_i, \theta) \big) \cdot Z(\theta')^K \cdot P(\theta)}$ \smallskip
        \State \textit{Conditionally Dependent:} \smallskip
        \State $\frac{P(\theta' \mid \mathcal{D})}{P(\theta \mid \mathcal{D})} \gets \frac{\exp \big( \beta \cdot \mathbf{R}(\mathcal{D}, \theta') \big) \cdot \mathbf{Z}(\theta) \cdot P(\theta')}{\exp \big( \beta \cdot \mathbf{R}(\mathcal{D}, \theta) \big) \cdot \mathbf{Z}(\theta') \cdot P(\theta)}$ \smallskip
        \If{$P(\theta' \mid \mathcal{D}) / P(\theta \mid \mathcal{D}) > \alpha\sim \mathcal{U}[0, 1]$}
        $\theta \gets \theta'$
        \EndIf
    \EndFor
    \State Return $\theta$
    \end{algorithmic}
\end{algorithm}

\p{MH Algorithm} We combine MH sampling with methods for approximating the normalizer in Algorithm~\ref{alg:norm}. At each iteration, we propose a new reward parameter $\theta'$. The robot then compares the probability of $\theta'$ with the probability of $\theta$, and accepts $\theta'$ with probability $\min \{1, P(\theta' \mid \mathcal{D}) / P(\theta \mid \mathcal{D})\}$. Similar to our analysis in Section~\ref{sec:norm}, any terms that do not depend on $\theta$ cancel out when we divide the posteriors. Each different approach for approximating the normalizer uses a different method for selecting $Z$ or $\mathbf{Z}$.
\begin{itemize}
    \item \textit{Ignore:} Set $Z(\theta) = 1$
    \item \textit{Sampling:} Approximate $Z(\theta)$ using \eq{M3}
    \item \textit{Maximum:} Approximate $Z(\theta)$ using \eq{M4}
\end{itemize}
These same equations extend to $\mathbf{Z}$, but now the robot samples from the space of datasets $\mathbb{D}$ instead of trajectories $\Xi$.

\vspace{-1em}
\subsection{Reward Learning with Double MH Sampling} \label{sec:A3}

In addition to the ignore, sample, and maximum methods from Section~\ref{sec:norm}, we can now introduce one final approach for approximating the normalizer. Standard MH approaches divide $P(\theta' \mid \mathcal{D})$ by $P(\theta \mid \mathcal{D})$ so that any terms that do not depend on $\theta$ are cancelled out. Here we take this concept one step further through \textit{double} MH sampling \cite{liang2010double}. At a high level, the double MH algorithm introduces an auxiliary variable such that, when we divide the posteriors, $Z(\theta)\cdot Z(\theta')$ appears in both the numerator and denominator, enabling us for the first time to cancel out the normalizing function. This shifts our problem: instead of approximating $Z(\theta)$, we need a method for generating the auxiliary variable.


\begin{algorithm}[t]
    \setstretch{1.0}
    \caption{Bayesian Reward Learning with Double MH}
    \label{alg:outer}
    \begin{algorithmic}[1]
    \State $\theta \gets$ sample from $P(\theta)$
    \For{each iteration}
        \State $\theta' \gets$ sample from $\Theta$ near $\theta$
        \State $\xi' \sim \mathcal{T}(\mathcal{D}, \theta')$ \Comment{inner sampler in Algorithm~\ref{alg:inner}}\smallskip
        \State $\frac{P(\theta' \mid \mathcal{D})}{P(\theta \mid \mathcal{D})} \gets \frac{\exp \beta \cdot \sum_{i = 1}^K \big(  R(\xi_i, \theta') - R(\xi', \theta') \big) \cdot P(\theta')}{\exp \beta \cdot \sum_{i = 1}^K \big(  R(\xi_i, \theta) - R(\xi', \theta) \big) \cdot P(\theta)}$\smallskip
        \If{$P(\theta' \mid \mathcal{D}) / P(\theta \mid \mathcal{D}) > \alpha \sim \mathcal{U}[0, 1]$}
        $\theta \gets \theta'$
        \EndIf
    \EndFor
    \State Return $\theta$
    \end{algorithmic}
\end{algorithm}

\begin{algorithm}[t]
    \setstretch{1.0}
    \caption{Inner Sampler for Double MH}
    \label{alg:inner}
    \begin{algorithmic}[1]
    \State Input dataset $\mathcal{D}$ and reward parameter $\theta$
    \State $\xi \gets$ sample trajectory from $\mathcal{D}$
    \For{each iteration}
        \State $\xi' \gets$ sample from $\Xi$ near $\xi$
        \If{$e^{\beta \big(R(\xi', \theta) - R(\xi, \theta)\big)} > \alpha \sim \mathcal{U}[0, 1]$}
        $\xi \gets \xi'$
        \EndIf
    \EndFor
    \State Return $\xi$
    \end{algorithmic}
\end{algorithm}

\p{Double MH Algorithm} We outline Double MH sampling for Bayesian reward learning in Algorithms~\ref{alg:outer} (outer sampler) and \ref{alg:inner} (inner sampler). For clarity we focus on \textit{conditionally independent} trajectories; it is straightforward to modify this pseudocode for the \textit{conditionally dependent} case. At each iteration, the outer sampler proposes a new $\theta'$. The inner sampler then inputs this $\theta'$ and generates a trajectory $\xi'$ from the distribution $P(\xi \mid \theta')$. The new trajectory --- which is sampled, and does not come from human demonstrations --- is the auxiliary variable. We leverage this auxiliary variable to cancel out the normalizing functions and avoid computing $Z(\theta)$. Specifically, the robot accepts $\theta'$ with probability:
\begin{equation} \label{eq:A6}
\min \bigg\{1, \frac{P(\theta') \cdot \big(\prod_{i = 1}^K P(\xi_i \mid \theta')P(\xi' \mid \theta)\big)}{P(\theta) \cdot \big(\prod_{i = 1}^K P(\xi_i \mid \theta)P(\xi' \mid \theta')\big)} \bigg\}
\end{equation}
Substituting in our human model and normalizing function:
\begin{equation*}
    \min \Bigg\{1, \frac{P(\theta') Z(\theta)^K Z(\theta')^K \cdot e^{\beta \sum_{i=1}^K \big(R(\xi_i, \theta')+R(\xi', \theta)\big)}}{P(\theta) Z(\theta')^K Z(\theta)^K \cdot e^{\beta \sum_{i=1}^K \big(R(\xi_i, \theta)+R(\xi', \theta')\big)}}\Bigg\}
\end{equation*}
Hence, the normalizing functions cancel out and we are left with the acceptance rule in Algorithm~\ref{alg:outer}. We note that this Double MH approach also extends to learning rewards from state-action pairs (instead of trajectories) if we replace $R(\xi, \theta)$ with the $Q$-function (i.e., the cost-to-go).

\p{Parameters} In our approach there are three main parameters for the designer to tune: a) the number of iterations in Algorithm~\ref{alg:outer}, b) the number of iterations in Algorithm~\ref{alg:inner}, and c) the rationality constant $\beta$. Increasing the number of outer and inner samples increases the expected accuracy of the learned $\theta$ \cite{liang2010double, park2018bayesian}, but also leads to longer run-times\footnote{We provide our code, environments, and additional results in \href{https://github.com/VT-Collab/Reward-Learning-with-Intractable-Normalizers}{this repository.} This includes an example of learning from state-action pairs.}. For example, when tested on our working example from Section~\ref{sec:norm}, Double MH took$~20 \%$ longer to complete the same number of MCMC iterations as sampling or maximization baselines.
\section{Simulations} \label{sec:sims}

Here we compare approaches for approximating the normalizer and performing Bayesian reward learning in controlled environments. We simulate noisily rational humans who provide multiple, conditionally independent demonstrations. The space of rewards $\Theta$ is continuous, and we attempt to infer the simulated human's $\theta \in \Theta$ based on their demonstrations.

\p{Independent Variables} We vary the robot's learning method across the algorithms introduced in Section~\ref{sec:mcmc}. This includes na\"ive robots that \textbf{Ignore} the normalizer, robots that approximate the normalizer using \textbf{Sampling} or \textbf{Maximum}, and our proposed \textbf{Double MH} approach. We also vary the simulated human's rationality $\beta$ at two levels: noisy humans ($\beta = 5$) and consistent humans ($\beta = 25$). These values of $\beta$ were identified through a preliminary round of simulations: below $\beta = 5$ humans acted almost completely randomly, and above $\beta = 25$ the humans converged to always select the optimal $\xi$.

\p{Dependent Variable} Each human samples their true reward $\theta$ uniformly at random. We report the \textit{Error} between the actual $\theta$ and $\hat{\theta}$, the mean of the robot's estimate: \textit{Error} $=\| \theta - \hat{\theta} \|$.

\begin{figure}[t]
	\begin{center}
		\includegraphics[width=0.8\columnwidth]{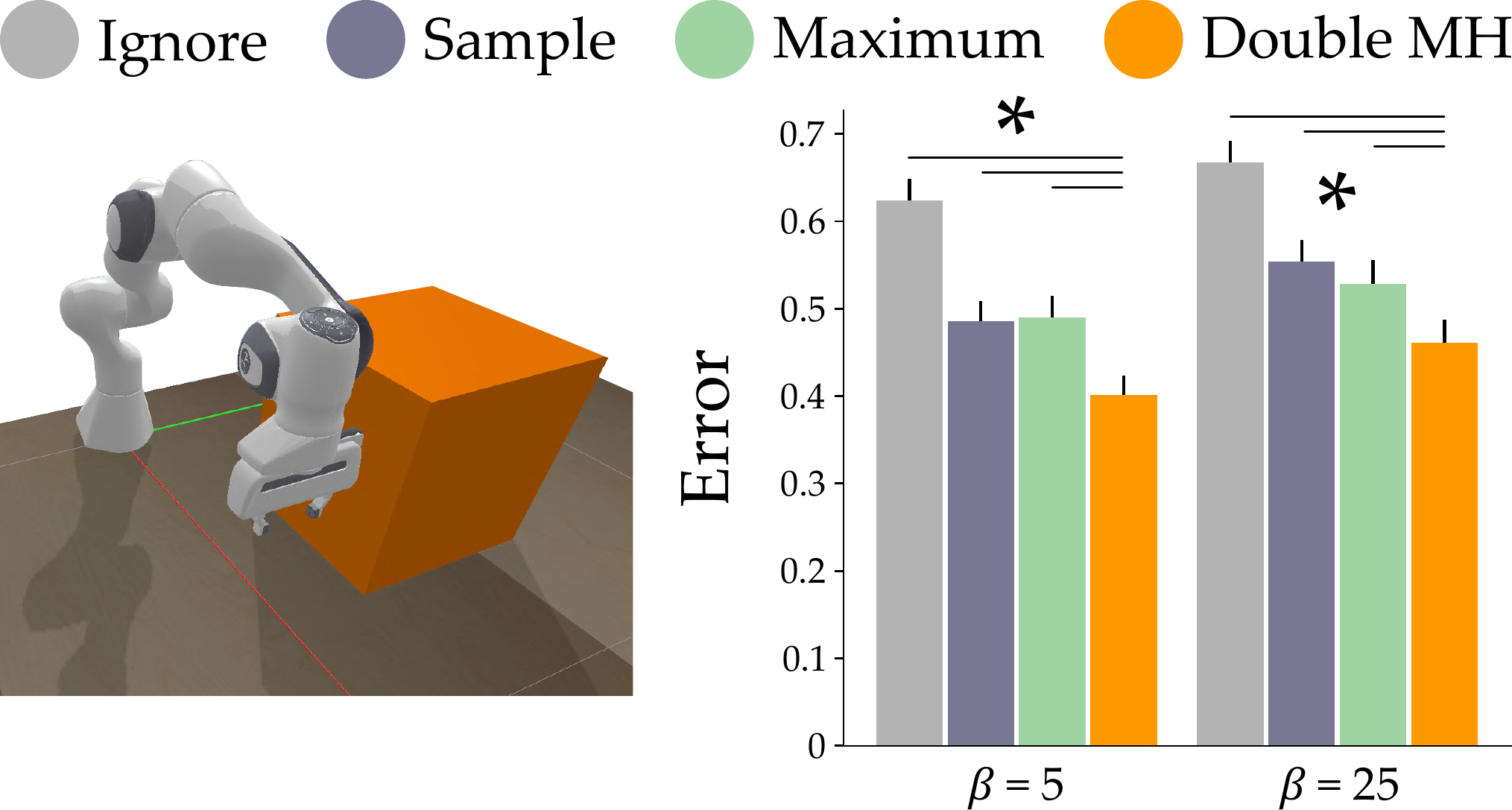}
		\caption{Results from the \textit{Push} simulation. (Left) The reward depends on the distance the box is moved and the distance the end-effector travels. (Right) Error in the learned $\theta$ across $100$ simulated humans. Error bars show standard error, and an $*$ denotes statistical significance ($p<.05$).}
		\label{fig:sim1}
	\end{center}
	\vspace{-1em}
\end{figure}

\begin{figure}[t]
	\begin{center}
		\includegraphics[width=0.8\columnwidth]{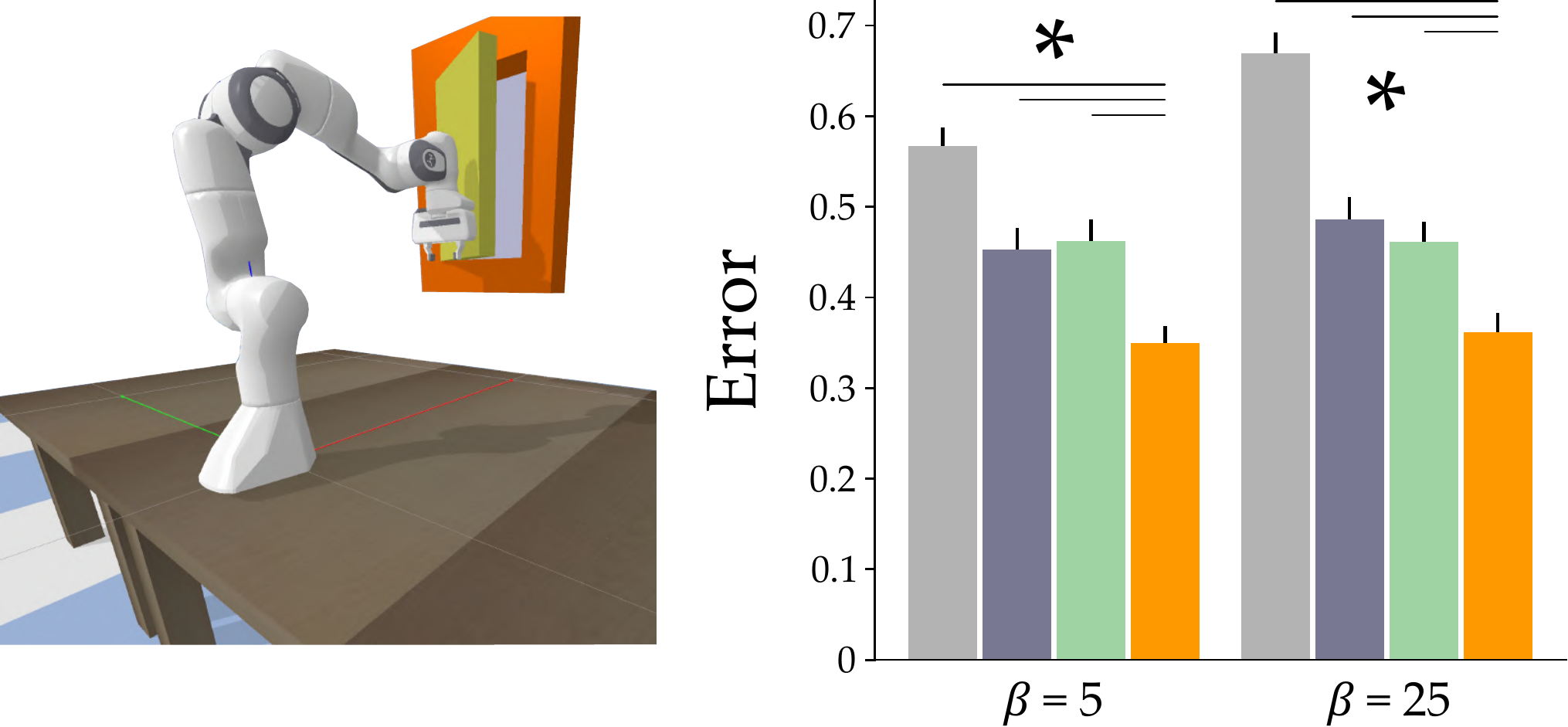}
		\caption{Results from the \textit{Close} simulation. (Left) The reward depends on the angle of the door and the robot's height from the table.}
		\label{fig:sim2}
	\end{center}
	\vspace{-2em}
\end{figure}

\begin{figure}[t]
	\begin{center}
		\includegraphics[width=0.8\columnwidth]{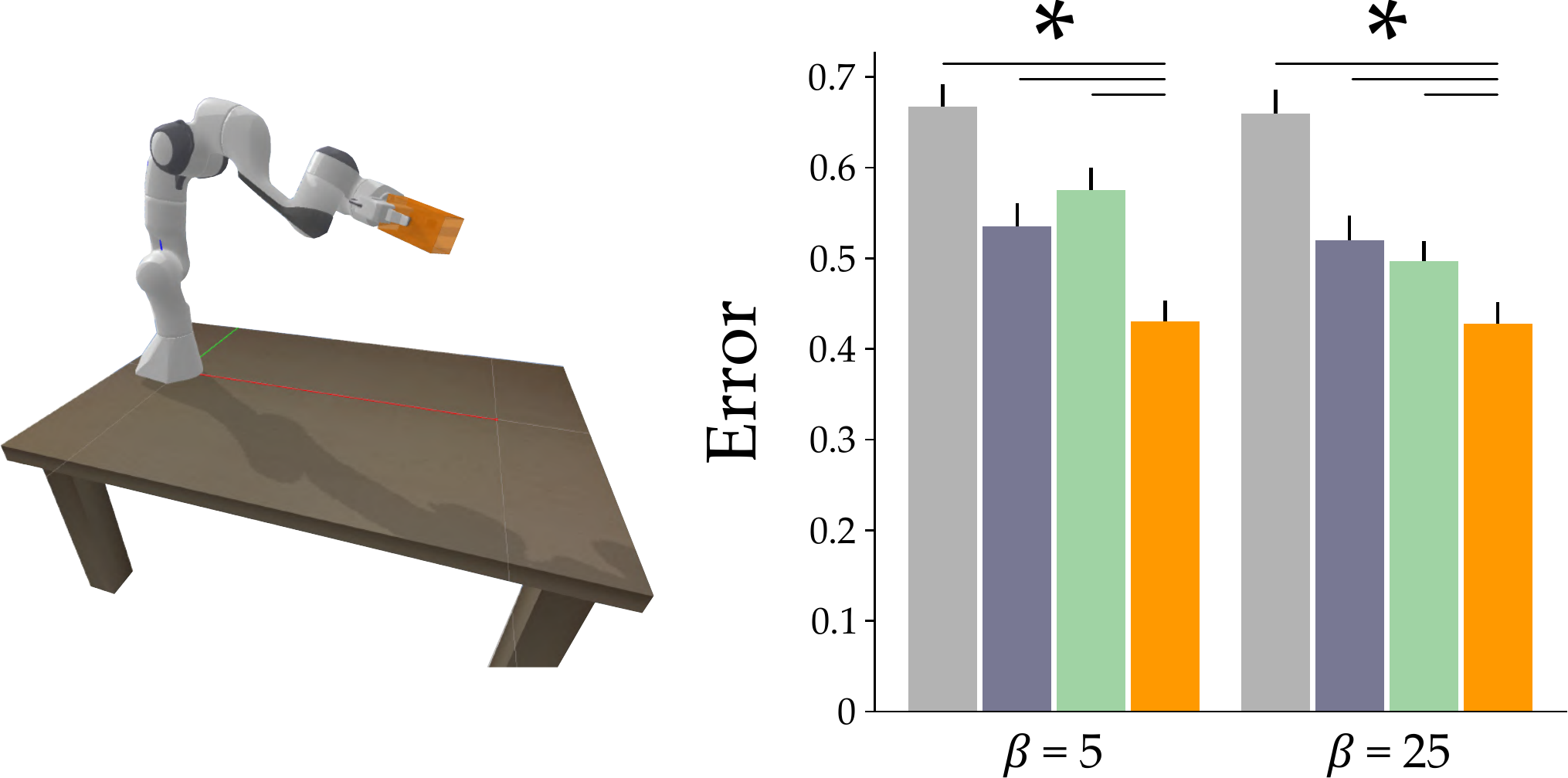}
		\caption{Results from the \textit{Pour} simulation. (Left) The reward depends on the orientation of the cup and the length of the robot's trajectory in joint space.}
		\label{fig:sim3}
	\end{center}
	\vspace{-2em}
\end{figure}

\p{Environments} We performed simulations across three dynamic physics-based environments where each environment had two features (see \ref{fig:sim1}, \ref{fig:sim2}, and \ref{fig:sim3}). In \textit{Push} the human's reward traded off between the distance the robot pushed the box and the length the robot travelled. In \textit{Close}, the human's reward traded off between pushing the door closed (i.e., the angle of the door) and keeping the robot's end-effector close to the table (i.e., the robot's height). Finally, in \textit{Pour}, the robot needed to pour coffee at a specific position, and the features were the distance travelled and holding the cup upright. In each environment, the robot had an accurate predictive model of the world and could simulate the outcomes of each trajectory.

\p{Procedure} Each simulated human chose a $\theta$ vector uniformly at random. The human then generated $K=3$ demonstrations of their desired motion so that $\mathcal{D} = \{\xi_1, \xi_2, \xi_3 \}$. These demonstrations were sampled from our noisily rational model in \eq{P2} and were conditionally independent (i.e., $\xi_2$ did not depend on $\xi_1$). The robot then observed $\mathcal{D}$ and used its Bayesian reward learning approach to get a mean estimate of $\theta$. For each environment, we repeated this procedure across $100$ simulated humans and reported the average results.

\begin{figure*}[t]
	\begin{center}
		\includegraphics[width=1.8\columnwidth]{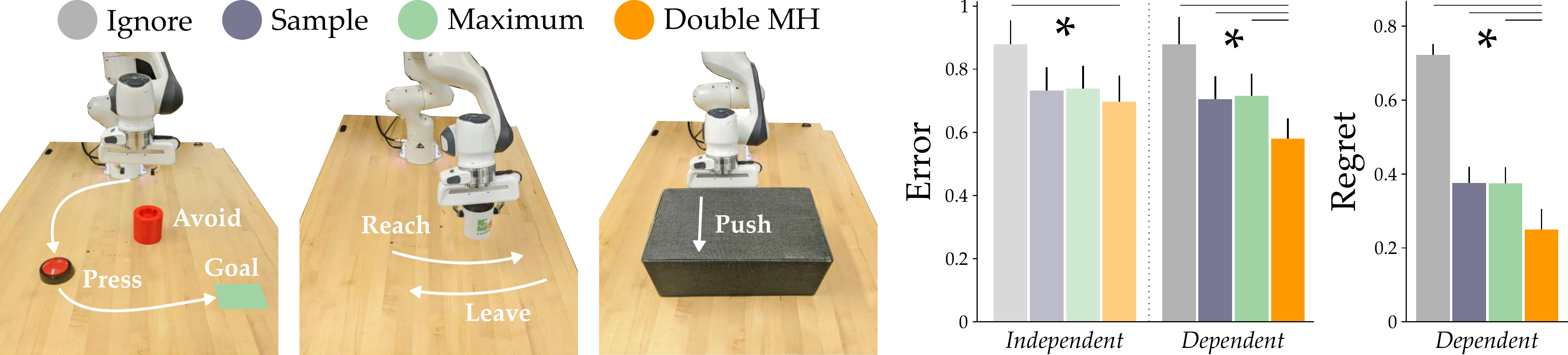}
		\caption{Results from our user study in Section~\ref{sec:user-study}. (Left) The \textit{Press}, \textit{Reach}, and \textit{Push} tasks. In each task, the robot moved along an initial trajectory, and users physically corrected the robot to teach it their desired behavior. (Right) \textit{Error} and \textit{Regret} averaged across the $10$ users and three tasks. Lower \textit{Regret} indicates that the robot's learned trajectory was better aligned with the desired behavior. Error bars show standard error, and an $*$ denotes $p<.001$.}
		\label{fig:user-study}
	\end{center}
	\vspace{-2em}
\end{figure*}

\p{Results} Our results for \textit{Push}, \textit{Close}, and \textit{Pour} are shown in Figures~\ref{fig:sim1}, \ref{fig:sim2}, and \ref{fig:sim3}. To analyze these results we first performed separate repeated measures ANOVAs on each environment and found that the normalizer approximation had a statistically significant effect. Post-hoc $t$-tests revealed that \textbf{Ignore} learned the \textit{least accurate} estimate (i.e., had the most error) across the board. At the other end of the spectrum, \textbf{Double MH} resulted in the \textit{most accurate} estimate for each environment and rationality $\beta$. Consider \fig{sim1} with $\beta = 5$ for instance: here $t$-tests show that \textbf{Double MH} has significantly lower error than \textbf{Ignore} ($t(99)=7.8$, $p<.001$), \textbf{Sample} ($t(99)=3.0$, $p<.05$), and \textbf{Maximum} ($t(99)=3.7$, $p<.001$). Overall, our simulation results in these three physics-based tasks suggest that (a) ignoring the normalizer altogether leads to inaccurate inference, and (b) using Double MH sampling to approximate the normalizer outperforms existing approximation methods.
\section{User Study} \label{sec:user-study}

We compared our proposed approach to existing approximations when learning rewards from actual users. In each task, the robot started with an initial trajectory and users physically \textit{corrected} the robot arm to better align its motion with their objective. To standardize these results, we first displayed the desired trajectory that the human should teach to the robot (i.e., we specified the user's reward parameters $\theta$). The participant's corrections were then used to infer an estimate of $\theta$, and we compared what the robot learned to the objective that the human was trying to teach the robot.

\p{Independent Variables} For this study, we varied the robot's learning along two factors: approximation type and conditional dependence. The robot used the \textbf{Ignore}, \textbf{Sample}, \textbf{Maximum}, and \textbf{Double MH} algorithms. We emphasize that \textbf{Ignore} \cite{cui2018active, brown2018risk, biyik2022learning}, \textbf{Sample} \cite{bobu2020quantifying, jonnavittula2021know, kalakrishnan2013learning}, and \textbf{Maximum} \cite{li2021learning,  levine2012continuous} come from prior work. We also compared \textit{Conditionally Independent} and \textit{Conditionally Dependent} versions of these algorithms. Recall from Section~\ref{sec:A1} that --- when the robot treats the human's inputs as conditionally dependent --- it recognizes that the human's current correction could build upon their prior corrections. Given that the human's corrections are sequential and interconnected, we anticipated that conditionally dependent learning would result in more accurate inference. To sample conditionally dependent corrections $\mathcal{D}'$ in \eq{A5}, we gave the robot an initial trajectory and then applied uniformly distributed perturbations to the waypoints along that trajectory.

\p{Dependent Measures} We recorded each participant's corrections and applied Bayesian reward learning to infer their objective $\theta$. As in Section~\ref{sec:sims}, we compared the \textit{Error} between the $\theta$ given to users and the robot's learned estimate $\hat{\theta}$: \textit{Error} $=\| \theta - \hat{\theta} \|$. We also computed the \textit{Regret} between the ideal trajectory $\xi$ (i.e., the trajectory we showed to participants which optimizes for $\theta$) and the learned trajectory $\hat{\xi}$ (i.e., the optimal trajectory under the robot's estimate $\hat{\theta}$).
\begin{equation} \label{eq:U1}
    \text{\textit{Regret}} = R(\xi, \theta) - R(\hat{\xi}, \theta), \quad \hat{\xi} = \text{arg}\max_{\xi \in \Xi} R(\xi, \hat{\theta})
\end{equation}
Lower \textit{Regret} means the robot has learned the correct behavior.

\p{Experimental Setup} Users taught the robot three tasks (see \fig{user-study}). One of these tasks (\textit{Push}) was consistent with the Simulations in Section~\ref{sec:sims}, and we introduced two new tasks to test the generality of our approach. In \textit{Press} the robot traded off between pressing a button, reaching a goal, and avoiding an obstacle. In \textit{Reach}, the robot tries to offer coffee to the user and then moves away after the coffee is delivered. Here \textit{Press} had four features and \textit{Push} and \textit{Reach} each had three features. For each task, the human provided three separate sets of corrections for three different values of $\theta$. Users first watched the robot demonstrate the ideal motion (i.e., the trajectory that optimized $\theta$) then gave a sequence of corrections to convey that $\theta$ to the robot.

\p{Participants} A total of $10$ participants from the Virginia Tech community took part in this study ($2$ female, ages $27 \pm 6.4$ years). Eight of the ten users had interacted with robots before, and the other two users had no prior experience in robotics. Users provided written consent under IRB$\#20$-$755$.

\p{Results} Our results averaged across these $10$ users and three tasks are summarized in \fig{user-study}. 

We first analyzed the effects of treating the human's corrections as conditionally independent or dependent. A repeated measures ANOVA revealed that conditionally dependent learning led to lower \textit{Error} across the board: $F(1, 29)=10.1$, $p < .001$. This result matched our intuition: it appeared that users often tried to fix something in their current correction based on what went wrong in the previous correction.

We next focused on the type of normalizer. Looking specifically at conditionally dependent learning, post-hoc analysis showed that \textbf{Double MH} resulted in lower \textit{Error} and \textit{Regret} as compared to each state-of-the-art alternative ($p<.001$). This suggests that --- not only does \textbf{Double MH} lead to a more accurate estimate of the human's reward --- but that estimate also results in robot trajectories that better match the human's desired behavior. See videos of our user study and the learned behaviors here: \url{https://youtu.be/EkmT3o5K5ko}
\section{Conclusion}

Our work is a step towards robot learners that infer the human's objective from demonstrations and corrections. In this paper, we explored the doubly-intractable nature of Bayesian reward learning, where the robot must reason over all possible trajectories and rewards. We grouped existing robotic approximations into three classes and theoretically derived their relative strengths and weaknesses. We then introduced a new Monte Carlo approximation method from the statistics community. Overall, our simulations and user studies suggest that this Double MH approach more accurately infers the human's objective, and is versatile enough to learn from independent demonstrations or interconnected corrections.


\balance
\bibliographystyle{IEEEtran}
\bibliography{IEEEabrv,bibtex}

\end{document}